# Reducing Bias in Deep Learning Optimization: The RSGDM Approach

1st Honglin Qin
Stevens Institute of Technology
Hoboken, USA

2nd Hongye Zheng
Chinese University of Hong Kong
Hong Kong, China

3rd Bingxing Wang
Illinois Institute of Technology
Chicago, USA

4th Zhizhong Wu
University of California, Berkeley
Berkeley, USA

5th Bingyao Liu*
University of California, Irvine
Irvine, USA

6th Yuanfang Yang
Southern Methodist University
Dallas, USA

*Abstract*—Currently, widely used first-order deep learning optimizers include non-adaptive learning rate optimizers and adaptive learning rate optimizers. The former is represented by SGDM (Stochastic Gradient Descent with Momentum), while the latter is represented by Adam. Both of these methods use exponential moving averages to estimate the overall gradient. However, estimating the overall gradient using exponential moving averages is biased and has a lag. This paper proposes an RSGDM algorithm based on differential correction. Our contributions are mainly threefold: 1) Analyze the bias and lag brought by the exponential moving average in the SGDM algorithm. 2) Use the differential estimation term to correct the bias and lag in the SGDM algorithm, proposing the RSGDM algorithm. 3) Experiments on the CIFAR datasets have proven that our RSGDM algorithm is superior to the SGDM algorithm in terms of convergence accuracy.

*Keywords-Deep Learning; First-order Optimizer; SGDM Algorithm; Differential Correction*

## I. INTRODUCTION

Deep neural networks have proven to be an exceptionally powerful tool across a wide range of applications, showcasing outstanding performance in numerous complex domains. In computer vision, for instance, they have driven substantial progress in image recognition, object detection, and scene understanding [1-3]. Similarly, in natural language processing, these networks have transformed tasks such as machine translation, sentiment analysis, and text generation [4-6]. Furthermore, in medical diagnostics, they have exhibited significant potential in fields like disease detection, imaging analysis, and personalized medicine [7-9]. Despite their success, as the number of layers in these networks increases, the training process becomes more complex and computationally demanding. This has led to a growing interest among researchers in developing and refining optimization techniques that can handle the challenges posed by deep networks. Among these techniques, Stochastic Gradient Descent (SGD) stands out as a particularly simple yet effective approach, widely utilized for addressing optimization problems in practical scenarios [10-12]. SGD works by updating the model parameters in the direction of the negative gradient of a differentiable loss function, with each update being based on a mini-batch of samples. The primary advantages of SGD include its fast training speed and the ability to achieve high accuracy.

However, one of the significant challenges with SGD is that it updates the model parameters with each mini-batch of samples. When there is substantial variation in the characteristics of different mini-batches, the direction of the parameter updates can change drastically. This can lead to difficulties in converging quickly to the optimal solution, as the updates may oscillate or follow a suboptimal path. To overcome this issue, researchers have developed a variety of SGD variants.

The first category includes approaches like Stochastic Gradient Descent with Momentum (SGDM) [13]. SGDM builds on the basic SGD algorithm by incorporating the first-order moment estimate of the gradient, which is achieved by calculating a moving average of the gradients of each mini-batch. This moving average helps to smooth out the updates, and effectively address the slow convergence problem often associated with standard SGD. As a result, SGDM has become a widely adopted method in the field of deep learning, and this paper aims to further improve upon this approach by addressing some of its limitations.

The second category consists of adaptive learning rate methods, which adjust the learning. These methods include popular algorithms such as AdaGrad [14], RMSProp [15], AdaDelta [16], and Adam [17]. Among these, Adam has gained significant popularity during training. However, while Adam often converges quickly on the training set, it tends to underperform in generalization compared to non-adaptive approaches like SGDM. This is particularly problematic when the model is applied to unseen data, where generalization is crucial.

To address the shortcomings of Adam, several improvements have been proposed, with AmsGrad [18] being one of the more notable attempts. AmsGrad modifies the Adam algorithm by preventing the adaptive learning rate from becoming too small, which in theory should help improve convergence accuracy. Nevertheless, recent studies have shown that AmsGrad does not significantly resolve the inherent limitations of adaptive optimization methods, and the actual improvements in performance are often marginal.

A more recent development in the field is the RAdam algorithm [19], which introduces a mechanism to rectify the changes in the adaptive learning rate. By adaptively correcting the learning rate adjustments, RAdam aims to provide a more stable and reliable training process, potentially combining the benefits of both adaptive and non-adaptive methods. This paper will explore these optimization techniques in depth, comparing their strengths and weaknesses, and proposing enhancements that could lead to better performance in training deep neural networks.

Many studies focus on the second category of adaptive learning rate methods, but they overlook the most fundamental issues. Both adaptive and non-adaptive methods use the exponential moving average method. These methods attempt to use the exponential moving average to approximate the gradient of the overall sample population. However, this method is biased and has lag. To address this issue, we propose the RSGDM algorithm. Our method calculates the difference in gradients, that is, the difference between the current iteration gradient and the gradient of the previous iteration, which is equivalent to the change in the gradient. We estimate the change in the current gradient using the exponential moving average at each iteration, and then sum this term with the estimated value of the current gradient, weighted by the current gradient. This approach can reduce bias and alleviate lag.

## II. Background

Stochastic Gradient Descent (SGD) is a cornerstone optimization algorithm widely employed in the training of deep learning models[20-21]. Originating from the broader field of optimization theory, SGD is designed to minimize a target function by iteratively moving toward the direction that most rapidly reduces the function's value. This approach not only makes the algorithm computationally efficient but also introduces noise that can help escape local minima, thus potentially leading to better generalization[22].

Despite its simplicity and effectiveness, SGD suffers from several limitations, such as slow convergence and sensitivity to the choice of the initial learning rate. Over the years, numerous variants of SGD, including SGD with momentum (SGDM), have been proposed to address these issues, enhancing the algorithm's performance and robustness in various training scenarios[23-25]. The development of more sophisticated optimizers, however, has not diminished the relevance of SGD, which remains a fundamental tool in the machine learning toolkit.

## III. Method

### A. RSGD Algorithm

We will illustrate the overall process of the RSGDM algorithm. Let $f(\theta)$ be a differentiable objective function with respect to $\theta$, and we aim to minimize the expected value of this objective function with respect to the parameters θ, i.e., minimize $E[f(\theta)]$. We use $f_1(\theta), f_2(\theta), \cdots, f_T(\theta)$ to represent the stochastic objective functions corresponding to time steps $1, 2, \cdots, T$ This randomness comes from the random sampling of each mini-batch or inherent noise in the function. The gradient is $g_t = \nabla_\theta f_t(\theta)$, which is the gradient vector of the objective function $f_t$ with respect to the parameters θ at time step $t$. We use $\Delta g_t = g_t - g_{t-1}$ to denote the difference in gradients between time step $t$ and time step $t-1$, which is the differential mentioned in this paper. Unlike the SGDM algorithm, which only performs exponential moving averages on $g_t$, the RSGDM algorithm performs exponential moving averages on both $g_t$ and $\Delta g_t$, and corrects the former using the moving average of the latter. The update formulas for the SGDM and RSGDM algorithms are listed below.

1. SGDM Algorithm:

$$m_t = \beta * m_{t-1} + (1-\beta) * g_t \quad (1)$$

$$\theta_t = \theta_{t-1} - \alpha * m_t \quad (2)$$

2. RSGDM Algorithm:

$$m_t = \beta * m_{t-1} + (1-\beta) * g_t \quad (3)$$

$$z_t = \beta * z_{t-1} + (1-\beta) * \Delta g_t \quad (4)$$

$$n_t = m_t + \beta * z_t \quad (5)$$

It can be seen that the RSGDM algorithm has additional formulas compared to the SGDM algorithm. Section 2 of this paper will prove that the estimate of $m_t$ in SGDM for the overall $g_t$ is biased, and we use formulas to correct this bias. The RSGDM algorithm does not introduce additional hyperparameters compared to the SGDM algorithm, which does not increase the burden of tuning parameters when training our model.

### B. Analysis of Bias and Lag

First, we analyze the bias in estimating the overall gradient using exponential moving averages. Starting from equation in the SGDM algorithm, we can derive:

$$m_t = (1-\beta) * \sum_{i=1}^{t} \beta^{t-i} * g_i \quad (6)$$

Taking the expectation of both sides of equation, we get:

$$E(m_t) = (1-\beta) * [\beta^{t-1} * E(g_1)$$
$$+ \beta^{t-2} * E(g_2) + \cdots + \beta * E(g_{t-1}) + E(g_t)] \quad (7)$$
$$= (1-\beta^t) * E(g_t) + (1-\beta) * \xi$$

It can be observed that $E(m_t) \neq E(g_t)$, where:

$$\xi = \sum_{i=1}^{t} \beta^{t-i}[E(g_i) - E(g_t)] = \beta^{t-1} * [g_1 - E(g_t)] \\ + \beta^{t-2} * [g_2 - E(g_t)] + \cdots + \beta * [g_{t-1} - E(g_t)] \quad (8)$$

When the number of iterations is large, $1 - \beta^t$ can be neglected, and the maximum bias arises from equation. If $g_t$ is a stationary sequence, i.e., $E(g_t) = C$ (where C is a constant), then $\xi = 0$, and at this point, $m_t$ is an unbiased estimate of $g_t$. However, in practical situations, this is obviously impossible, so $m_t$ is a biased estimate of $g_t$, and the bias primarily comes from $\xi$. Moreover, this bias leads to lag. For example, if the gradient is consistently increasing, the smaller gradient values from previous historical moments will also cause the estimated value $m_t$ to be somewhat smaller. Or if the gradient has been increasing but starts to decrease at a certain moment, the estimated value $m_t$ may not have reacted yet due to the influence of historical gradients. This is the impact of lag as discussed in this paper. To address this situation, we propose the RSGDM algorithm, which uses the differential (change) of the gradient to correct the estimated value $m_t$. Intuitively, it can be understood in this way: if the gradient is increasing and the differential estimate is also greater than 0, then this correction term plays a role in accelerating convergence. If the gradient is increasing and at a certain moment begins to decrease, this correction term will play a role in adjusting the direction of gradient descent. Below, we explain the advantages of the RSGDM algorithm from a formulaic perspective: we can derive:

$$z_t = (1 - \beta) * \sum_{i=2}^{t} \beta^{t-i} * \Delta g_i \quad (9)$$

By taking the expectation on both sides of equation in the RSGDM algorithm, we can obtain:
$$E(n_t) = (1 - \beta^t) * E(g_t) + (1 - \beta) * \xi + \beta(1 - \beta) \\ * \left( \sum_{i=2}^{t} \beta^{t-i} * \Delta g_i \right) \quad (10)$$

$n_t$ is the corrected estimate of $g_t$. We can let $\varsigma = \xi + \beta(\sum_{i=2}^{t} \beta^{t-i} * \Delta g_i)$. It is not difficult to see that $\varsigma$ is the main source of bias in the RSGDM algorithm. We expand $\varsigma$ to get:

$$\varsigma = \sum_{i=1}^{t-2} \beta^{t-i} * [g_{i+1} - E(g_t)] = \beta^{t-1} * [g_2 - E(g_t)] \\ + \beta^{t-2} * [g_3 - E(g_t)] + \cdots + \beta^2 * [g_{t-1} - E(g_t)] \quad (11)$$

We compare the bias term $\xi$ of the SGDM algorithm with the bias term $\xi$ of the RSGDM algorithm. It can be seen that $\xi$ is influenced by the historical gradients $g_1, g_2, \cdots, g_{t-1}$, while $\varsigma$ is influenced by the historical gradients $g_2, g_3, \cdots, g_{t-1}$. Since t is large and $\beta$ is less than 1, $\beta^{t-1} * [g_1 - E(g_t)]$ approaches 0 and can be neglected, then we can obtain $S = \beta * \xi$. It can be seen that the bias term $\varsigma$ of the RSGDM algorithm lacks the influence of the historical gradient $g_1$, and because $\beta$ is less than 1, $|\varsigma| \leq |\xi|$. In summary, we can conclude that the RSGDM algorithm has a smaller bias compared to the SGDM algorithm and is less influenced by historical gradients (mitigating lag).

## IV. EXPERIMENT

### A. Experiment Settings

In this section, we demonstrate through experiments that our RSGDM algorithm has more advantages than the SGDM algorithm. We set $\beta = 0.9$ and $\alpha = 0.01$. To validate the algorithm's superiority, we conducted experiments on image classification tasks with the CIFAR-10 and CIFAR-100 datasets [26]. Both the CIFAR-10 and CIFAR-100 datasets comprise RGB images with a resolution of 32×32, featuring a training set of 50,000 images and a test set of 10,000 images. We perform classification of 10 categories on the CIFAR-10 dataset and classification of 100 categories on the CIFAR-100 dataset.

We use the ResNet18 model and the ResNet50 model for image classification tasks on the CIFAR-10 and CIFAR-100 datasets, comparing each task using the SGDM algorithm and our RSGDM algorithm, with the evaluation metric being classification accuracy. We use the PyTorch deep learning framework, and the hardware environment for training is a single NVIDIA RTX 2080Ti GPU. The batch size in the experiment is set to 128, and the two hyperparameters for the SGDM algorithm and the RSGDM algorithm are set the same, including momentum and initial learning rate. During training, we used weight decay to prevent overfitting, with the decay parameter set to 5×10−4 5×10−4, and the learning rate is halved every 50 epochs.

### B. Experiment Results Analysis

Table 1 Experiment Result on CIFAR-10

| Method | TrainSet | ValidSet |
|---|---|---|
| SGDM | 1 | 0.9948 |
| RSGDM | 1 | 0.9462 |

Table 2 Experiment Result on CIFAR-100

| Method | TrainSet | ValidSet |
|---|---|---|
| SGDM | 0.9998 | 0.7670 |
| RSGDM | 0.9998 | 0.7727 |

Table 1 and Table 2 respectively present the accuracy rates of image classification on the CIFAR-10 and CIFAR-100 datasets using different optimizers with ResNet18 and ResNet50. We can observe that on the CIFAR-10 dataset, both SGDM and RSGDM achieved a training accuracy of 100%, and in terms of test accuracy, our RSGDM outperformed SGDM by 0.14%. On the CIFAR-100 dataset, the training accuracy for both SGDM and RSGDM was 99.98%, and in test accuracy, RSGDM exceeded SGDM by 0.57%.

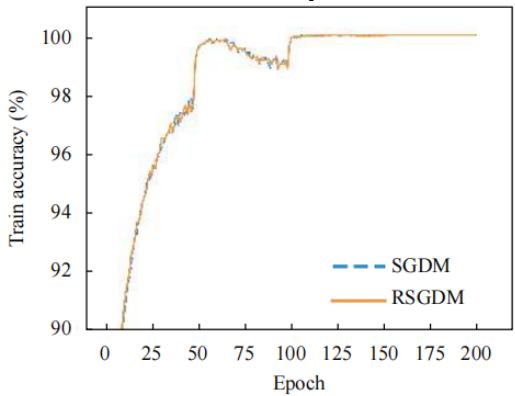

Figure 1. The performance of the ResNet18 model with the CIFAR-10 dataset

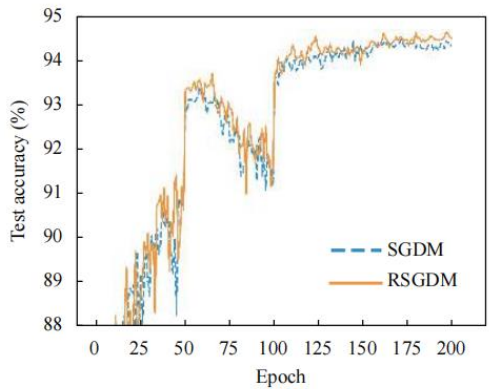

Figure 2. ResNet18's testing accuracy with CIFAR-10

Figures 1 to 2 display the experimental results of ResNet18 on CIFAR-10 using the SGDM and RSGDM algorithms, including training accuracy, training loss, test accuracy, and test loss. Since our experimental setup involves halving the learning rate every 50 epochs, it is evident that there are fluctuations in all four graphs at epochs 50, 100, and 150. Overall, in terms of training accuracy and training loss, both methods are largely the same in terms of convergence speed and convergence accuracy, but in the later stages of convergence, our RSGDM algorithm has an advantage.

Figures 3 to 4 present the experimental results of ResNet50 on CIFAR-100 using the RSGDM and SGDM algorithms. Similar results to CIFAR-10 can be drawn, where both methods are almost identical in terms of training accuracy and training loss, but in terms of convergence accuracy, the RSGDM algorithm significantly outperforms the SGDM algorithm on this dataset. As can be seen from the test accuracy graph, after 100 epochs, our RSGDM method consistently maintains a higher accuracy than SGDM, and the final accuracy is 0.57% higher than SGDM. This further demonstrates the effectiveness of our method.

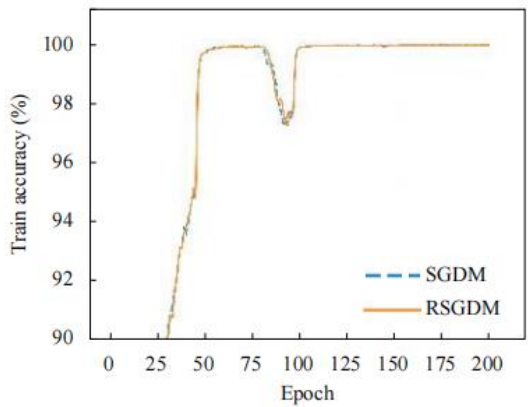

Figure 3. ResNet18's testing accuracy with CIFAR-100.

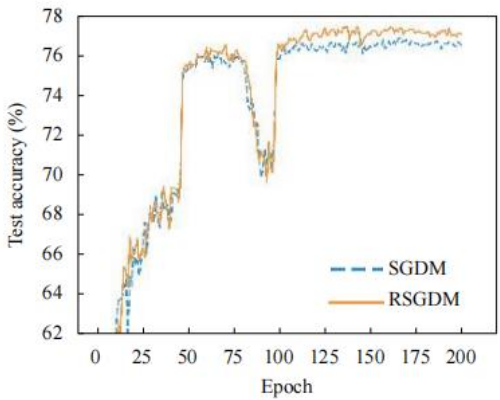

Figure 4. ResNet18's testing accuracy with CIFAR-100

V. CONCLUSION

The research presented in this paper introduces the RSGDM algorithm, an innovative approach that significantly ameliorates the inherent bias and lag associated with traditional SGDM gradient estimation methodologies. By integrating a differential correction term that dynamically adjusts based on the differences between consecutive gradients, RSGDM not only addresses the primary deficiencies of SGDM but also enhances the overall robustness of the learning process.

Empirical evaluations conducted using the CIFAR-10 and CIFAR-100 datasets have substantiated the superior performance of RSGDM, demonstrating its enhanced convergence properties and accuracy in comparison to the conventional SGDM method. These results not only reinforce the validity of RSGDM as a potent optimization tool but also highlight its potential to facilitate more effective training of deep neural networks, particularly in applications demanding high precision and reliability. Looking forward, the RSGDM algorithm opens new avenues for further research and development. Its adaptable framework makes it a promising candidate for exploration in other complex machine learning tasks beyond image recognition, such as time series analysis and unsupervised learning. Additionally, the principles underlying the differential correction strategy employed in

RSGDM may inspire novel optimization algorithms that could further refine the efficiency and accuracy of training deep learning models. In conclusion, RSGDM represents a pivotal step forward in the optimization of deep neural networks. By mitigating the limitations of gradient estimation that have long challenged traditional methods, it sets a new benchmark for the development of advanced optimization algorithms in the field of deep learning. Future work will focus on extending the applicability of RSGDM to a broader range of datasets and problem domains, potentially revolutionizing the way we approach challenges in artificial intelligence research.